\documentclass[final]{article}

\usepackage[nonatbib]{nips_2016}

\usepackage{nips_2016}

\usepackage[utf8]{inputenc} 
\usepackage[T1]{fontenc}    
\usepackage{url}            
\usepackage{booktabs}       
\usepackage{amsfonts}       
\usepackage{nicefrac}       
\usepackage{microtype}      
\usepackage{amsmath,amssymb,subcaption,graphicx}
\usepackage{color}

\title{TreeView: Peeking into Deep Neural Networks Via Feature-Space Partitioning}

\author{
  Jayaraman J. Thiagarajan\thanks{This work was performed under the auspices of the U.S. Dept. of Energy by Lawrence Livermore National Labora- tory under Contract DE-AC52-07NA27344.} \\ \bf Bhavya Kailkhura \\
  Center for Applied Scientific Computing\\
  Lawrence Livermore National Laboratory\\
  Livermore, CA 94550 \\
  \texttt{\{jjayaram,kailkhura1\}@llnl.gov} \\
   \And
  Prasanna Sattigeri \\ \bf Karthikeyan Natesan Ramamurthy \\
  Mathematical Sciences Department\\
  IBM Thomas J. Watson Research Center \\
  Yorktown Heights NY 10598 \\
  \texttt{\{psattig,knatesa\}@us.ibm.com} \\
}

\begin{document}

\maketitle

\begin{abstract}
With the advent of highly predictive but opaque deep learning models, it has become more
important than ever to understand and explain the predictions of such models. Existing approaches
define interpretability as the inverse of complexity and achieve interpretability at the cost of accuracy. This introduces a risk of producing interpretable but misleading explanations. As humans, we are prone to engage in this kind of behavior~\cite{mythos}.
In this paper, we take a step in the direction of tackling the problem of interpretability without compromising the model accuracy. We propose to build a Treeview representation of the complex model via hierarchical partitioning of the feature space, which reveals the iterative rejection of unlikely class labels until the correct association is predicted. 
\end{abstract}

\section{Interpretability in Machine Learning}
Analysis and inference from rich, multivariate data is a ubiquitous problem in science and engineering. The quality of inference depends on the quality of relevant features extracted, a.k.a. data representation. While recent deep learning approaches have revolutionized representation learning, there is an increase in the number of researchers that complain about the difficulty in interpreting the decisions of such complex models. Despite their remarkable capacity to learn complex functions from data, there is a fair amount of distrust in applying such models particularly to domains such as  as medicine, criminal justice, and finance, where humans are directly impacted. To complicate matters further, the term interpretability itself does not have a clear definition \cite{mythos}, and can be understood at various levels. This poses a real problem in the universal acceptance of machine learning solutions, and deserves to be tackled head on.


The simplest, yet a popular, definition of interpretability is as the inverse of complexity - e.g., there is a long standing argument that linear models are more interpretable than non-linear models because of their simplicity in the parameter space. There have been deliberate efforts to develop models
that are interpretable in the recent past~\cite{rivest1987learning},~\cite{malioutov2013exact},~\cite{ustun2014methods} .  Also, the recently developed
locally interpretable model-agnostic explanations (LIME) approach poses interpretation as a sparse local fitting problem~\cite{ribeiro2016model}. However, a simple model is less expressive by definition and hence one cannot express a complex process such as human cognition using a simple model with guaranteed generalization and high accuracy. One of the fundamental problems seems to be that the language for interpreting machine learning models has not matured, and it still relies on incomplete ideas such as simplicity. Furthermore, drawing parallels with human decision making, a complex decision will take a lot of words for a satisfactory explanation as opposed to a simple decision, and the explanation will inevitably involve abstractions which will be revealed on a need-to-know basis. This paper attempts to take a step in the direction of a developing a tool for interpreting complex models, the \emph{TreeView}, without comprising its performance. In particular, we propose to explore deep learning models using sequential elimination via feature-space partitions. This is essentially a process of understanding via hierarchical partitioning and association of feature space where the most undesirable options are discarded and the scope of options is progressively narrowed down until exactly one option is left.





\section{Proposed Approach}
A typical supervised learning algorithm consists of two stages. The first stage is feature learning or extraction where the input data space $\mathcal{X}$ is transformed into a representation or feature space $\mathcal{Y}$. The second stage is predictive learning which maps the feature space $\mathcal{Y}$ to the label space $\mathcal{Z}$. For example, $\mathcal{Y}$ could amplify the factors that are discriminatory across classes thereby aiding the predictive inference. The transformations learned in these two stages can be concisely denoted as $T_{XY}: \mathcal{X} \rightarrow \mathcal{Y}$ and $T_{YZ}: \mathcal{Y} \rightarrow \mathcal{Z}$, whereas the full map is given as $T_{XZ} = T_{XY} \circ T_{YZ}$. The representation space $\mathcal{Y}$ that is important for the high performance of the learning task, is also opaque and un-interpretable if the feature learning algorithm is complex and non-linear. Our goal is to provide interpretable explanations for this space using the proposed TreeView model.

\begin{figure}[t]
\centering
\begin{center}
\begin{subfigure}[t]{1.\linewidth}
\centerline{\includegraphics[width=0.5\linewidth]{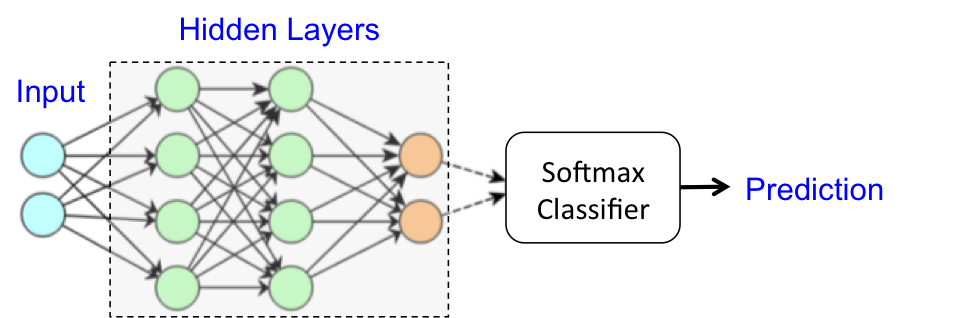}}
\end{subfigure}

\vspace{0.2in}

\begin{subfigure}[t]{1.\linewidth}
\centerline{\includegraphics[width=0.8\linewidth]{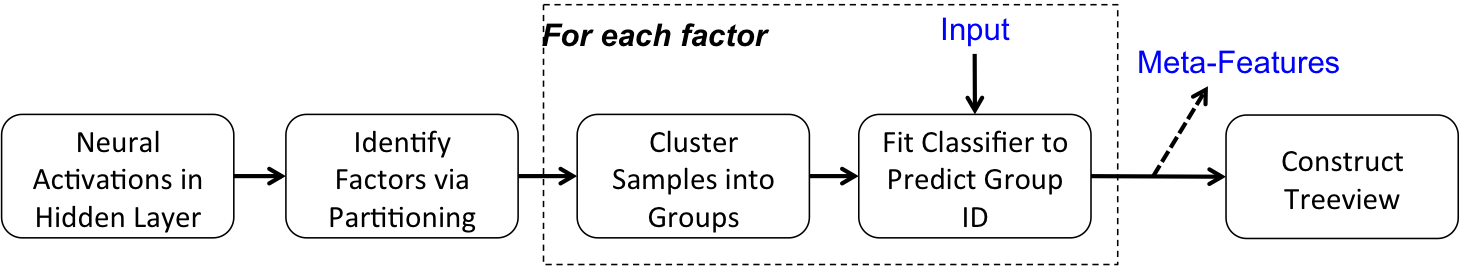}}
\end{subfigure}

\end{center}
\caption{(Top) The hidden representation of even a simple fully connected deep neural network can be highly opaque for interpretation. (Bottom) Proposed approach for building \textit{TreeView} visualizations using a surrogate model.}
\label{fig:overview}
\end{figure}

In this section, we describe the proposed approach in the context of understanding features learned by a deep neural network. We consider only fully connected networks although it is possible to conceive of extensions to other architectures. Figure \ref{fig:overview}(a) illustrates a simple fully connected deep network with a softmax layer for class prediction. Unraveling the mechanics of the hidden layer representations can provide interesting insights into the trained model. This is a main distinction of our approach compared to other existing methods that attempt to learn a simpler surrogate (e.g. linear models) to explain the predictions for individual examples.
 

Figure \ref{fig:overview}(b) shows an overview of our approach. We decompose the feature space $\mathcal{Y}$ into $K$ potentially overlapping factors. Let us denote the feature dimensions corresponding to these factors using the sets $\{S_1, \ldots, S_K\}$. We obtain $K$ different clusterings of samples, one for each of the subspaces  of $\mathcal{Y}$ given by $S_i$. We then construct a $K-$dimensional \emph{meta-feature} for each sample as a collection of its $K$ cluster labels. The TreeView framework uses these meta-features used in a decision tree to create an easily interpretable visualization of the mechanics of the learned deep network.

\subsection{Identifying Factors}
The hidden neurons in a deep neural network learn different aspects of the training data samples that facilitate the task. The same aspect of the training data can be captured by different neurons because of the distributed nature of learned representations. Hence it is reasonable to expect that this feature space can be clustered into factors, each of which represents one aspect of the input space at some level of abstraction. We elaborate this by extending the notation provided before. For layer $\ell$, the activations are given by the matrix $\mathbf{Y}_{\ell} \in \mathbb{R}^{N_{\ell} \times T}$, where $N_{\ell}$ and $T$ are the number of filters and training samples respectively. We cluster $\mathbf{Y}_{\ell}$ into $K$ factors, $\{\mathbf{F}^{\ell}_i\}_{i=1}^K$, such that each factor is comprised of a set of hidden neurons that have similar distribution of activations across the whole training set. 


\subsection{Constructing Meta-Features}
The second important step is to create meta-features that are easier to interpret, but still perform well in the learning task. We construct the surrogate model as follows: First, we consider the activations in each factor $i$, $\mathbf{F}^{\ell}_i \in \mathbb{R}^{N^{\ell}_i \times T}$, and cluster the $T$ samples in this factor into $L$ groups. We then create a meta-feature matrix $\mathbf{M} \in \mathbb{R}^{K \times T}$ by aggregating the $K$ cluster labels for each sample. Subsequently, we train a predictor $\mathcal{P}^{\ell}_i$ that directly predicts the cluster label for a sample using the input space examples. In our case, the predictor used is a random forest classifier. This will enable the analyst to understand each factor using the input examples directly, and circumvent the need to train an approximate model (e.g. linear). A decision tree surrogate is finally created for the neural network classifier using the meta-features.

\subsection{TreeView Design}
\begin{figure}[t]
  \centering
  \centerline{\includegraphics[width=0.85 \linewidth]{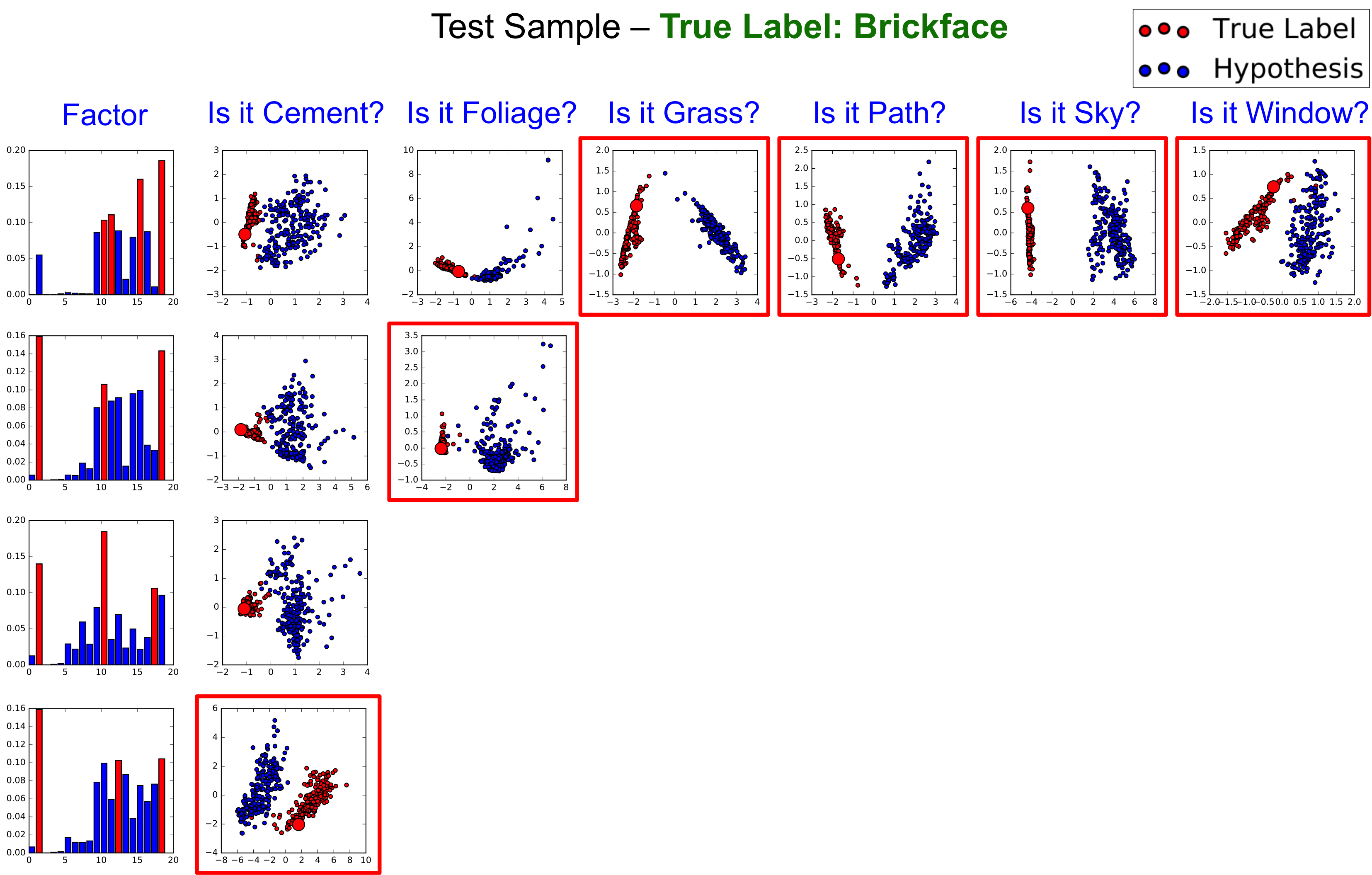}}
%
\caption{\textit{Treeview} visualization for a sample which is correctly classified by the neural network.}
\label{pos}
\end{figure}

Consider a test example whose prediction we want to interpret using the \textit{Treeview} visualization. Assuming that its true label is known, we compute the meta-feature using the predictors corresponding to each of the factors. This meta-feature is used with the decision tree surrogate to predict the label and the sequence of nodes visited in the decision tree during this prediction is traced. As shown in Figure \ref{pos}, each column in a \textit{TreeView} corresponds to the query if the sample belongs to a specific class, while the rows indicate the sequence of factors required to effectively reject the hypotheses and predict the correct label. We consider a hypothesis to be rejected if the activations corresponding to the factor ($\mathbf{F}^{\ell}_i$) clearly discriminates the true class and the hypothesis. In addition, the relative ranks of the input space features in the factor-specific predictor $\mathcal{P}^{\ell}_i$ are shown to allow the user to create a mental map between the class labels and input data.

\begin{figure}[t]
  \centering
  \centerline{\includegraphics[width=0.9 \linewidth]{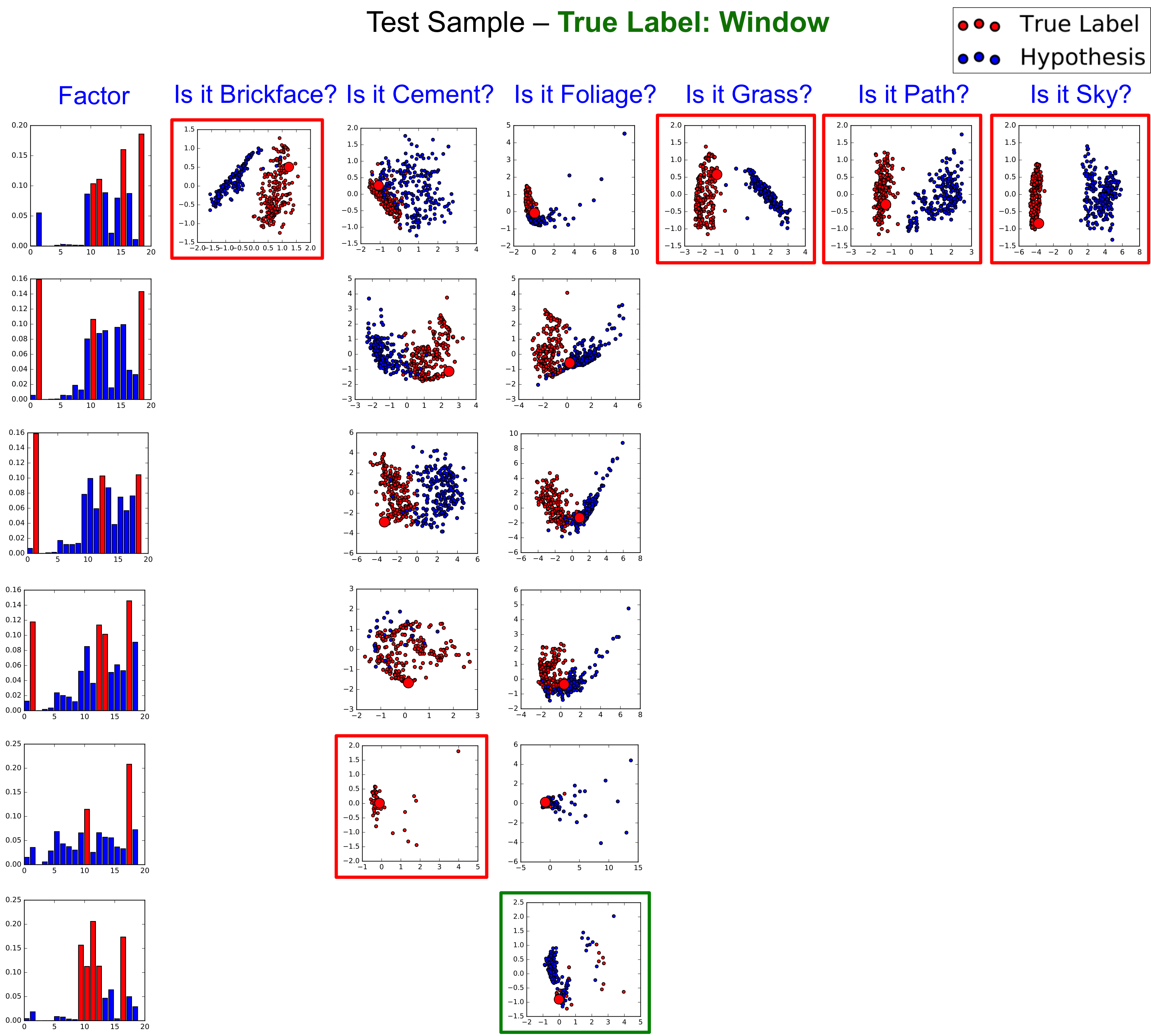}}
\caption{\textit{Treeview} visualization for a sample which is wrongly classified by the neural network.}
\label{neg}
\end{figure}

\section{Demonstration}
We demonstrate the usage and effectiveness of the proposed approach using the UCI Image segmentation dataset. The dataset contains $2310$ instances, each of which belong to one of the $7$ outdoor image categories. Each instance is a $3\times3$ region that were hand segmented and is characterized by $19$ attributes that show describe either the semantic or statistical properties of the instance. The deep neural network used for classification consisted of $3$-hidden layers with sizes $128$-$64$-$64$. At each hidden layer the dropout rate was fixed at $0.1$. Random forests were used to predict the meta-feature for each of the factors which in turn were fed into a decision tree classifier. As described earlier, the proposed surrogate model does not compromise accuracy for simplicity, which can be evidenced by the classification accuracies ($96\%$ and $94\%$ respectively). Figure \ref{pos} shows the TreeView visualization for a correctly classified example. While the root factor rejects the hypothesis for the classes \textit{Grass, Path, Sky, Window} (indicated by a red bounded box) right away, a sequence of three more factors were required to reject the hypotheses for \textit{Cement}. A fine-grained analysis of the high-ranked features from the input space (marked in red) can provide more insights into the connection between the input and label spaces, thereby enabling the analyst to validate their mental map between the spaces. In contrary, the negative example in Figure \ref{neg} illustrates a case where the factors identified by the network were unable to clearly discriminate between \textit{Foliage} and the true class label \textit{Window}. The \textit{Treeview} visualization allows a convenient transition between factors, class labels, and the input data space, while staying relevant to the features inferred using the neural network.

\bibliographystyle{plain}
\bibliography{ref}

\begin{thebibliography}{1}

\bibitem{mythos}
Zachary~C Lipton.
\newblock The mythos of model interpretability.
\newblock {\em IEEE Spectrum}, 2016.

\bibitem{malioutov2013exact}
Dmitry Malioutov and Kush Varshney.
\newblock Exact rule learning via boolean compressed sensing.
\newblock In {\em Proceedings of the 30th international conference on machine
  learning}, pages 765--773, 2013.

\bibitem{ribeiro2016model}
Marco~Tulio Ribeiro, Sameer Singh, and Carlos Guestrin.
\newblock Model-agnostic interpretability of machine learning.
\newblock {\em arXiv preprint arXiv:1606.05386}, 2016.

\bibitem{rivest1987learning}
Ronald~L Rivest.
\newblock Learning decision lists.
\newblock {\em Machine learning}, 2(3):229--246, 1987.

\bibitem{ustun2014methods}
Berk Ustun and Cynthia Rudin.
\newblock Methods and models for interpretable linear classification.
\newblock {\em arXiv preprint arXiv:1405.4047}, 2014.

\end{thebibliography}
\end{document}